# Reconstructing hourly traffic on every street of a capital from a sparse counting network: a graph neural network applied to Paris

Xavier Jurado[1,2*], Cédric Wemmert[2] & Nicolas Reiminger[3**]

[1] *AIR&D, F, 67000, Strasbourg, France*

[2] *University of Strasbourg, CNRS, ICube, UMR7357, F, 67000, Strasbourg, France*

[3] *University of Strasbourg, CNRS, ENGEES, ICube, UMR7357, F, 67000, Strasbourg, France*

* Corresponding author #1: Tel. +33 (0)6 15 11 22 54, Mail. xjurado@air-d.fr

** Corresponding author #2: Tel. +33 (0)6 31 26 75 88, Mail. nicolas.reiminger@engees.unistra.fr

## Abstract

Micro-scale air quality models require hourly traffic data on every road segment, whereas permanent counters cover only a small part of the network. A graph neural network (GNN) was developed to reconstruct hourly traffic flows on the 32,163 road segments of Paris intra-muros from the 1,573 segments equipped with a counter (4.9%). The model adapts the Graph Neural Network for Urban Interpolation (GNNUI) to a real, sparse counting network. Flows are decoded on the edges of the junction graph (one edge per segment, both directions summed), a traffic climatology serves as an offset in a zero-inflated negative-binomial likelihood, and training proceeds by masking counters. Eight-fold spatial cross-validation, with counters sharing a published series merged before splitting, distinguishes interpolation at a withheld counter in an instrumented district ($R^2 = 0.89$, error of a quarter of the mean flow) from extrapolation to a district without counters ($R^2 = 0.74$, a third of the mean flow). These scores hold on main axes where counters are located, while relative errors are about 50% on the few counted local streets. Ablations show that the climatology offset tends to help extrapolation. Road attributes, time cycles and the live state of the counting network matter most, while weather inputs have no significant effect. The masking rate is the only inherited hyperparameter with a clear effect. When evaluated on January–August 2026 (outside the training period), the frozen model stayed around the error band of previous years.

## 1. Introduction

Road networks structure the spatial layout of modern cities, yet many of the environmental pressures they experience are closely linked to the traffic they carry. Road traffic is, for instance, the largest source of nitrogen oxides ($NO_x$) emissions in European cities (Thunis, 2018), as are ultrafine particles (Trinh et al., 2025), and a major determinant of their concentrations at the street scale, alongside the strong influence of urban geometry (Reiminger et al., 2024) and local climatology (Reiminger et al., 2025). Consequently, micro-scale air quality models require traffic volumes for each road segment, from which emissions are estimated using information on the vehicle fleet and vehicle speeds (Lejri et al., 2018). Thus, uncertainty regarding traffic volumes or, worse still, a complete lack of traffic data, has a direct impact on emissions estimates and, ultimately, on simulated pollutant concentrations (Martín et al., 2024, 2025). Accurate traffic information is also required for a broader range of urban applications, including noise mapping, population exposure assessment, and the evaluation of traffic-management measures such as low-emission zones (Mavzeroğlu et al., 2026). When these applications rely on hourly meteorological data, average daily or annual traffic volumes are insufficient. Indeed, the temporal variability of traffic must be considered so that emissions are attributed to the appropriate time periods, allowing high-resolution hourly modeling approaches to reach their full potential (Jurado & Reiminger, 2026). However, traffic observations are generally available only at a limited number of locations. Permanent traffic counters, such as inductive loops, are typically installed along the main road axes and cover only a small fraction of the road network, leaving traffic volumes on most road segments unobserved. These missing flows must therefore be reconstructed, potentially at an hourly resolution, from the observations available at instrumented locations. This study addresses the reconstruction of hourly traffic volumes on an urban road network from incomplete observations. Thus, the aim is not to forecast future traffic, but to estimate, hour by hour, the traffic on every road of the city, including the many segments where no measurement is available. Such estimates are needed, for instance, as traffic input to street-level emission and air quality models.

Three families of methods can provide such estimates. The first is traffic assignment, the standard approach for estimating the flow on every link of a network. In the four-step model, a travel demand is distributed over the routes of the network (Ortúzar & Willumsen, 2011), while agent-based and microscopic simulators such as MATSim and SUMO represent individual trips and give time-resolved outputs (Horni et al., 2016; Lopez et al., 2018). Because they link flows to demand and travel behavior, these models are the natural tool for scenario analysis. However, they require demand data, a detailed description of the network and a calibration step. Moreover, fitting the model to the observed counts constrains the flows at the counting sites, but does not guarantee their accuracy on the roads without counters, which are precisely those of interest here. The second family relies on probe vehicles, i.e., GPS-equipped vehicles and navigation devices. Their data are widely used to estimate speeds and travel times (Herrera et al., 2010), and can also provide traffic volumes if the share of vehicles they represent is known. Eisinga & Lorkowski (2025) combine probe data with fixed counts to model this share, or penetration rate, at the road level and estimate volumes over the whole network. This approach requires the penetration rate to be estimated for each road and period, and some probe datasets cannot be shared, which prevents independent reproduction. Floating car data is thus a useful option where they are available, but their restricted access motivates approaches based only on open data. The third family estimates the traffic on a road without counters from the traffic observed elsewhere, the road characteristics and its surroundings. At the annual scale, Sfyridis & Agnolucci (2020) combine clustering and regression to estimate the average annual daily traffic for several road classes, and Ma et al. (2026) add contextual and spatial features and assess their model with a spatial validation. These methods cover the whole network, but annual values do not capture the hourly variations required in the context of the present work. At finer time scales, estimating traffic at unmonitored locations from simultaneous measurements is a spatiotemporal kriging problem. Graph neural networks (GNNs) are well suited to this problem, as they learn dependencies along the road graph and combine them with temporal information. However, most GNN studies on traffic deal with forecasting at monitored locations (Jiang & Luo, 2022; Jin et al., 2024), and the reference benchmarks concern speed forecasting on motorway and major-road sensor networks (Li et al., 2018; Yu et al., 2018).

Estimating traffic at locations without any past observation is a different problem, which some models address. Kriging convolutional networks combine local observations and graph convolutions for spatial interpolation, without modeling temporal dynamics (Appleby et al., 2020). Inductive Graph Neural Network Kriging (IGNNK) is trained on random subgraphs in which some observations are masked, so that it can be applied to new nodes and new graphs (Wu, Zhuang, Labbe, et al., 2021). Later models describe spatial and temporal dependencies in more detail: SATCN combines several spatial aggregation functions with temporal convolutions (Wu, Zhuang, Lei, et al., 2021), and INCREASE accounts for spatial proximity, functional similarity and transitions between locations, the latter being derived from taxi trajectories (Zheng et al., 2023).

The Graph Neural Network for Urban Interpolation (GNNUI) extends the masking approach of IGNNK to urban street networks, using street attributes as node features and a loss function adapted to counts with many zeros (Kaiser et al., 2026). It was evaluated on daily cycling volumes in Berlin and hourly taxi volumes in Manhattan, with a simulated sensor coverage ranging from 90% to 1%. To our knowledge, it is the closest model to the problem addressed in the present work, and it is the starting point of this study. However, its evaluation does not reflect the conditions of a citywide traffic estimate. First, the benchmarks describe a single mode, cycling or taxi trips, reconstructed from trajectories that cover all the streets of the network, and the sensors are selected at random among them. In contrast, emission models require total motorized flow, which is measured by loop detectors located mainly on main roads, rarely on local streets and not at all in some districts. The sensor coverage is therefore not only sparse but also unevenly distributed, which random sampling does not reproduce. Second, the test set is a fixed random 10% of the segments, which does not distinguish interpolation between nearby sensors from extrapolation to a district without any. Finally, the flow level is learned from the node features and the neighboring nodes only, so that its transfer to a district without counters is not tested.

This study therefore adapts GNNUI to the real counting network of Paris, where only 4.9% of the road segments are equipped with counters, using four years of hourly loop counts. Only open data are used: the road network of the national mapping agency, the counts and roadworks published by the city, the observations of the national meteorological service, open land use, population and public-transport layers and the public calendar. The reconstruction can thus be reproduced by anyone, without commercial trajectory data or travel surveys.

Three adaptations address the limitations identified above. First, flows are estimated on the road segments themselves (the edges of the junction graph), which are the units used by emission models, so that each segment keeps its own attributes and its own estimate (total flow, both directions summed on two-way segments). Second, a median hourly profile per road class, computed from the training counters, is used as a reference level, and the network learns a correction relative to this reference rather than the absolute flow. Third, the model is evaluated with a spatial cross-validation designed for counting networks, which distinguishes two cases: estimating a counter inside an equipped area and estimating a whole area without counters. Within this framework, the study examines which design choices and inputs matter most, including the settings inherited from GNNUI, and whether the model trained on 2022–2025 remains reliable outside this period. Finally, the study provides four years of hourly traffic flows on every segment of the Paris road network.

Section 2 describes the study area, the data, the model and the evaluation method. Section 3 presents the results which are discussed in Section 4 together with the limitations and future work. Section 5 concludes.

## 2. Material and methods

### *2.1. Study area and data availability*

The study was conducted over Paris intra-muros, France, a dense built-up area of about 105 km$^2$ with a dense permanent traffic couting network. All the data used in this study are open data and can be summarized as follows.

- Road network: the BD TOPO® database of the French national mapping agency (IGN), version 3.5, edition of 15 December 2025, which provides the geometry and the attributes of every road segment (see Section 2.2).

- Traffic counts: the raw hourly counts of the permanent counting network of the City of Paris (electromagnetic loops), published as open data, from January 2022 to August 2026 (see Section 2.3).
- Meteorological data: hourly temperature, precipitation and wind speed monitored at the Paris-Montsouris station of Météo-France, retrieved throught Meteosat.
- Calendar and events: public and school holidays, transit-strike days and the roadworks published by the city.
- Land use, population and public transport: OpenStreetMap, INSEE gridded population and Île-de-France Mobilités stops and lines (see Suppelmentary Table S1).

It should be noted that no commercial data, such as floating car data, and no travel demand survey were used, so that the whole reconstruction can be reproduced from public sources.

### *2.2. Road network*

The road network used in this study is the junction graph of the IGN BD TOPO® database (version 3.5, edition of 15 December 2025). Junctions are the nodes of the graph and road segments are its directed edges, with one edge per BD TOPO segment, so that the output of the model is produced on the same objects as the emissions of any air quality model. The graph extends several kilometers beyond the ring road (in order to give the model some context on both sides of the city limit) and comprises 44,763 nodes and 61,993 edges, of which 22,475 nodes and 32,163 edges are inside the commune of Paris. All the statistics given hereafter concern the intra-muros segments and counters (inside the surrounding ring road). Each segment carries its BD TOPO attributes (importance class, number of lanes, carriageway width, regulatory speed, length, one-way, urban, bridge and tunnel flags) and descriptors derived from the graph (betweenness centrality, degree of the end junctions, distance to the center and to the nearest major road, local road density, sinuosity, bearing), completed by land use, gridded population and the proximity of public transport (see Supplementary Table S1). Figure 1 shows the graph by importance class, which is the attribute that organizes both the network and its counters. The three upper classes form a sparse skeleton of axes, while the local streets of class 5 represent half of the segments.

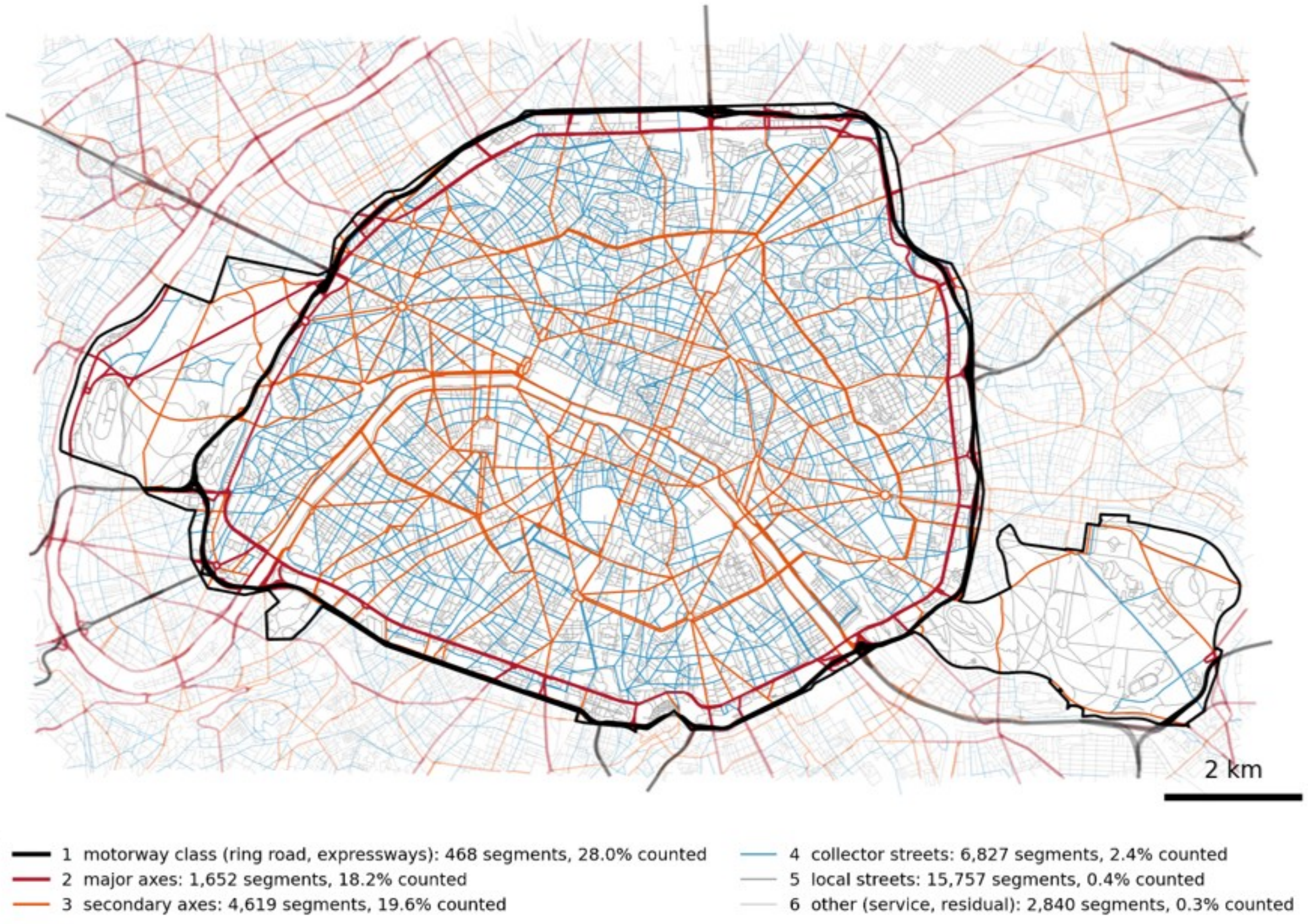

**Figure 1**. Road network from the BD TOPO database colored by importance class. Segments within Paris intra-muros are shown in full color, and those of the surrounding buffer zone included in the graph in lighter color. The thin black line marks the city boundary. For each class, the legend gives the number of segments within Paris and the percentage equipped with a counter.

### *2.3. Traffic count data*

The raw traffic data come from the counting network of the City of Paris, which consists of electromagnetic loops and is published as open data as hourly flows per counting point (one series per road section, which several counting points can share; see Section 2.7). Each point measures one direction of a carriageway. The points were mapped to the BD TOPO segments by geometry and name, and the two directions of a carriageway were summed, since a line source emits the traffic of both directions. Three rules follow from this:

(1) Hours in which only part of the channels of a carriageway report are discarded, because the stored total would be a fraction of the real flow.

(2) Two-way segments where a single direction is counted (241 segments) keep their flow as an input of the model but are never used as a training target or as a test value.

(3) Cells imputed by the data provider (26% of the record, flagged as such) are treated in the same way.

After this processing, the dataset contains 1,579 instrumented segments (1,573 intra-muros, i.e. 4.9% of the 32,163 segments) and 35,061 hourly snapshots over 2022–2025, 328 of the segments being two-way carriageways with both directions counted. It is worth noting that the counting network is structurally biased (see Supplementary Table S5). Importance classes 1 to 3 (motorways, major and secondary axes) are covered at 18–28%, whereas classes 4 to 6, which represent 79% of the edges and the residential streets, are covered below 2.5%. The problem is therefore both to interpolate between nearby sensors on the main axes and to extrapolate to districts and road classes without any counter. Half of the unmonitored intra-muros segments are within three junctions of a counter and nine out of ten within ten.

Finally, since flow is conserved along a chain of segments without junction, the segments of such a chain inherit the counter of the chain as virtual training targets (371 additional segments). They follow the fold of their parent counter, are never used in the evaluation, and are hidden from the inputs whenever their parent is a test counter.

### *2.4. Model inputs*

The model takes four families of inputs for each hourly snapshot, detailed in Supplementary Table S1:

(1) A static description of each edge, with 28 features comprising the BD TOPO attributes and graph descriptors presented above, land use, population density and the proximity of public transport.
(2) A description of each node at the current hour, with 7 features giving its position and degree and the observed flows on the unmasked sensor edges adjacent to the junction.
(3) The dynamic state of each edge, with 4 features giving the observed flow and loop occupancy where a counter reports and flags indicating whether the edge is masked or missing.
(4) A temporal description of the hour with 36 features: 33 are shared by all edges (cyclic encodings of the hour, day and month, calendar flags, the weather observed at Paris-Montsouris, the live mean flow of the counting network with its lags and trends and 3 are defined per edge (roadworks, transit strikes and daylight flags).

All continuous features are standardized. It should be noted that at every hour the model sees the flows measured on the unmasked sensors, both locally and citywide, so that the reconstruction is conditioned on the observed state of the network at that hour and not only on the calendar and the weather.

### *2.5. Deep learning model*

The model reconstructs hourly traffic flows on road segments by combining observations from the surrounding network with information about the target segment and the hour being reconstructed. Junctions form the nodes of the graph, whereas the quantities to be predicted are attached to its edges, the road segments. The architecture therefore first builds a representation of each junction and its neighborhood, then uses the representations of a segment's two end junctions to predict its flow. This separates the sharing of information across the network from the use of characteristics specific to each segment. A segment without a counter can consequently receive information from observed locations through the graph while retaining its own physical and temporal context.

The model used in this study, illustrated in Figure 2, with a size of about $3.7 \times 10^5$ parameters, is adapted from GNNUI (Kaiser et al., 2026), which itself follows IGNNK, the inductive graph neural network for kriging of Wu, Zhuang, Lei, et al. (2021). The differences with GNNUI are listed in Supplementary Table S2. One forward pass processes one hour on the whole graph. Figure 3 presents the same model as a sequence of steps on the network, from a segment described by its attributes and bounded by two junctions, to the junctions gathering what the counters read in their surroundings, the hour and the state of the network modulating this information, a prior level given by the climatology of similar roads, and the non-linear layers that combine all of it into a flow distribution.

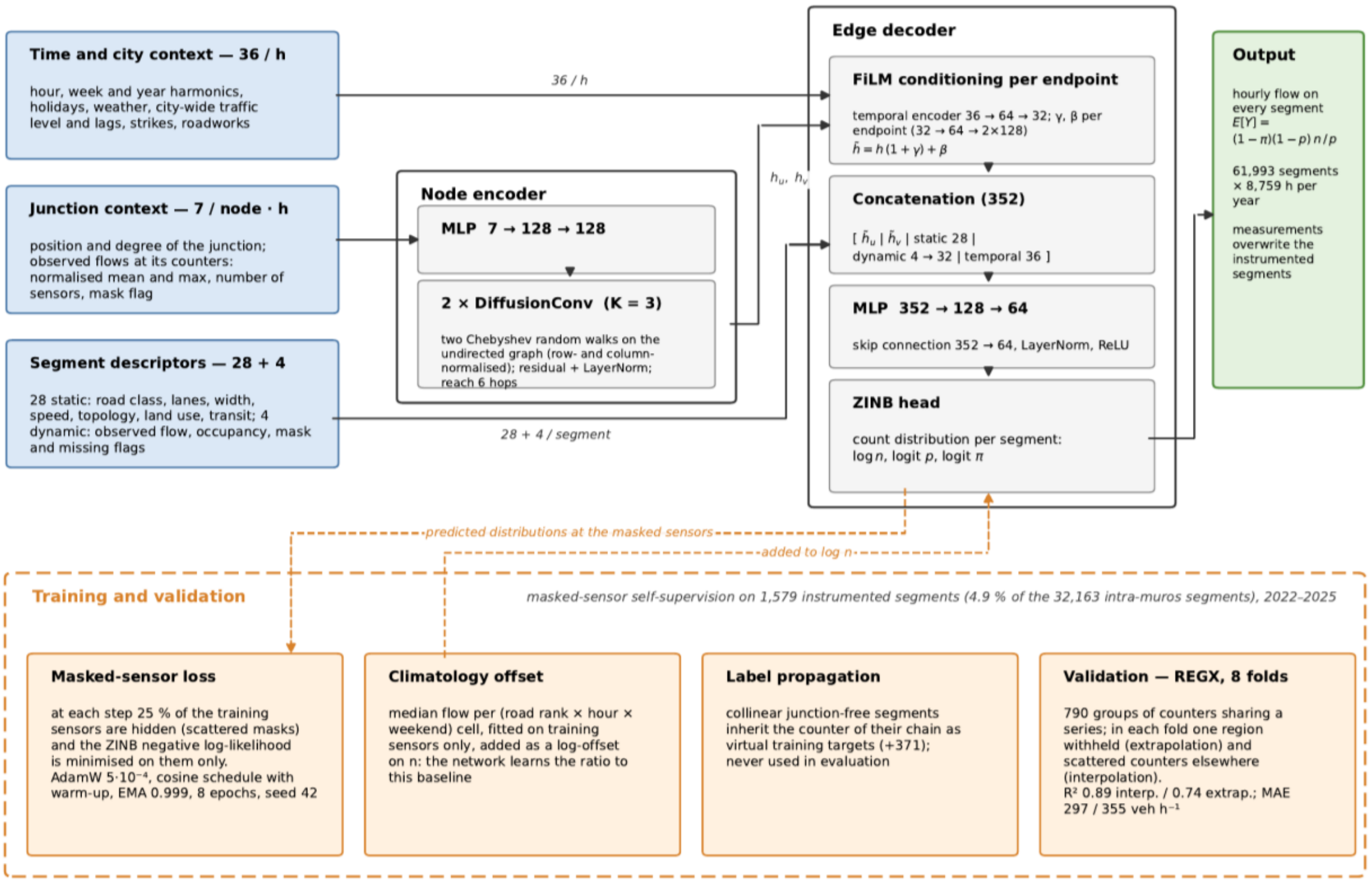

**Figure 2**. Traffic reconstruction model. Top: one forward pass per hour on the junction graph, from the inputs (left) to the ZINB flow distribution on every segment (right); $h_u$, $h_v$ are the embeddings of the two end junctions of a segment and $\tilde{h}_u$, $\tilde{h}_v$ the same embeddings after FiLM conditioning by the hour; layer sizes are those of the deployed model (367,459 parameters). Bottom: training choices and validation protocol (dashed arrows show where each acts).

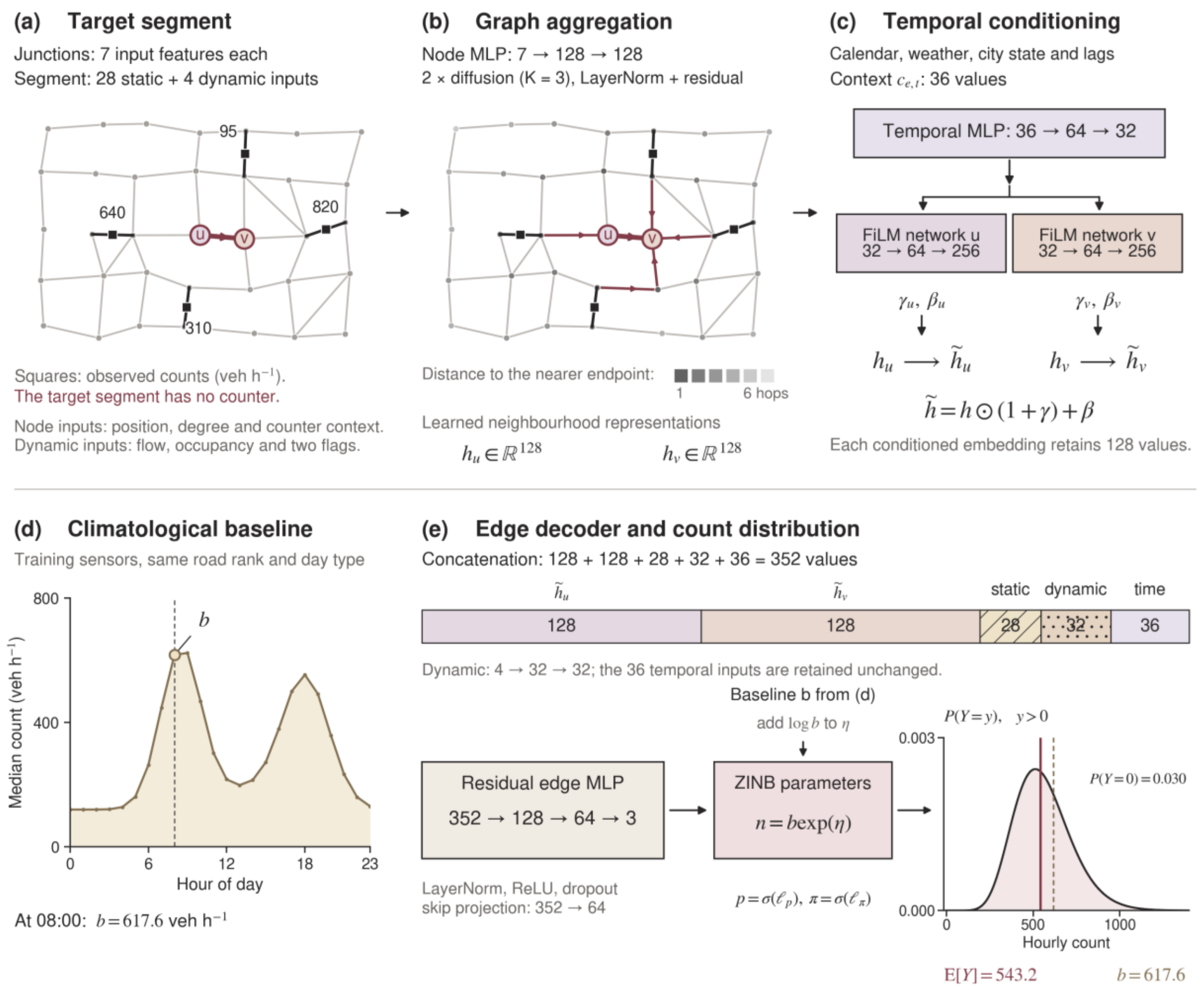

**Figure 3**. Schematic overview of the model. The network topology, counts and curves are illustrative, but the dimensions correspond to those of the model used in this study. (a) The target segment $e=(u, v)$ has no counter. It is described by 28 static and 4 dynamic features, and each of its two junctions by 7 features. (b) Two diffusion layers aggregate information from up to six hops around each junction, so that the surrounding counters contribute to the embeddings $h_u$ and $h_v$ of the junctions $u$ and $v$. (c) The 36 temporal features of the current hour are encoded and used to rescale these embeddings through two FiLM networks. (d) The median count of the segment's cell (road rank × hour × weekday or weekend), computed from the training counters, gives a prior level $b$. (e) The resulting vector of 352 values is passed to a residual edge

decoder. The decoder predicts the three parameters of a zero-inflated negative binomial distribution of the hourly count, with $log(b)$ included as an offset.

The node features are first embedded by a two-layer perceptron of 7, 128, 128 neurons, then propagated over the graph by diffusion convolution (Li et al., 2018). For this propagation, the junction graph is symmetrized with each segment contributing to an edge in both directions, so that the message passing does not depend on the direction of travel (the latter only enters the model through the edge decoder described below). The diffusion uses two normalizations of the adjacency matrix $A$ of this undirected graph, $A_f=D^{-1}A$ and $A_b=AD^{-1}$ with $D$ the degree matrix, which weights the contributions by the degree of either the receiving junction or its neighbor (see Supplementary Section S6). One layer computes Eq. 1:

$$h'=ReLU\big(\Theta\big[h \parallel T_1(A_f)h \parallel \cdots \parallel T_K(A_f)h \parallel T_1(A_b)h \parallel \cdots \parallel T_K(A_b)h\big]\big) \qquad (1)$$

where $T_k$ is the Chebyshev polynomial of order $k$ in the random-walk matrix and $\Theta$ a linear map from $(2K+1) \times 128$ to $128$. Each layer is followed by a normalization and a residual connection. The deployed model stacks two such layers of order $K=3$, which gives a receptive field of six hops around each junction. The order is inherited from the model family, whereas the depth was reduced from three to two layers; both are tested in Section 3.6.

The 36 temporal features are encoded by a perceptron of 36, 64 and 32 neurons into a summary of the hour, which modulates the embedding of each end junction of an edge by feature-wise linear modulation (FiLM; (Perez et al., 2018)), $\tilde{h}=h(1+\gamma)+\beta$, with separate $(\gamma, \beta)$ networks for the upstream and downstream junctions of the segment. The same junction embedding can thus be read differently at the morning peak and at night and differently depending on whether the junction is entered or left by the segment, which is where the direction of travel enters the model, together with the one-way flag of the static descriptors.

The two modulated junction embeddings are then concatenated with the static descriptors $s$ of the segment, their dynamic features encoded to 32 dimensions, $d$, and the temporal features $c$, kept as they are. A two-layer perceptron with hidden sizes 128 and 64, a skip connection from the input and layer normalization maps $z$ to the three parameters of a zero-inflated negative-binomial (ZINB) distribution (Lambert, 1992), namely $log(n)$, $logit(p)$ and $logit(\pi)$. Decoding at the edge rather than at the node keeps the per-segment descriptors intact, which matters when two carriageways of the same junction carry very different flows.

The hourly flow $y$ of a segment is modeled as Eq. 2 and Eq. 3:

$$P(y)=\pi 1[y=0]+(1\text{-}\pi)NB(y;n,p) \qquad (2)$$

$$E[Y]=(1\text{-}\pi)\frac{n(1\text{-}p)}{p} \qquad (3)$$

where $\pi$ is the weight of an additional point mass at zero, which accounts for closed or empty segments (the negative-binomial component can also produce zero counts), and the negative binomial carries the over-dispersion of hourly counts. The reported flow is the expectation of this distribution.

A climatology offset is then added to the logarithm of the size parameter. With $b$ the median flow of the cell (road rank × hour of the day × weekday or weekend) to which the segment belongs, computed on the training sensors only, the first output of the decoder $\eta$ is shifted as $log(n)=\eta+log(b)$, as in a negative-binomial regression with offset (McCullagh & Nelder, 1989). The expected flow is then the baseline multiplied by a learned correction as given in Eq. 4:

$$E[Y]=b\cdot(1\text{-}\pi)e^{\eta}\frac{1\text{-}p}{p} \qquad (4)$$

The network learns how far the segment departs from the typical road of its cell rather than an absolute level. The intent is that this correction transfers to districts never seen while the level comes from the baseline, which is tested by ablation in Section 3.3. The baseline is recomputed on the training sensors of each fold and on all sensors for deployment, so that no held-out flow ever enters it; its construction is detailed in Supplementary Section S7.

### *2.6. Training*

The model is trained by masking the counters (He et al., 2022; Wu, Zhuang, Labbe, et al., 2021). At each training step, 25% of the counters drawn randomly over the network are hidden. Their observations are removed from the inputs (node, dynamic and citywide features) and the ZINB negative log-likelihood is minimized on these hidden counters only. The model thus learns, at every hour, to reconstruct a counter it cannot see from the counters it can see and from the descriptors of the segment, which is close to the task at deployment, where no counter is hidden and the targets have never been counted. Hiding whole districts instead of scattered counters was also tested (see Section 3.5). The optimization uses AdamW (Loshchilov & Hutter, 2019) with a learning rate of $5 \times 10^{-4}$, a cosine schedule with warm-up, an exponential moving average of the weights, mixed precision and gradient clipping, for up to 8 epochs, the number of epochs being selected on the training counters (see Supplementary Table S4). One cross-validation fold trains in one to three and a half hours on a single GTX 1080 Ti GPU.

### *2.7. Evaluation methodology*

The performance of the model is evaluated with an eight-fold spatial cross-validation designed to avoid two pitfalls of counting networks. Withholding whole regions corresponds to the block cross-validation recommended for spatially structured data (Roberts et al., 2017), so that the validation is aligned with the task at deployment, which is the estimation of streets and districts the model has never seen.

The first possible leak comes from the counters sharing a series. The city publishes one flow series per road section, so that several counting points can carry the same series, identical on their measured cells, and a spatial hold-out that separates them would place the exact twin of a test counter in the training set. Counters sharing a series are therefore merged into 790 indivisible groups before the split, and a group is scored through one representative counter.

The second is that a random hold-out only tests interpolation between nearby counters, whereas traffic levels vary between districts. The 790 groups are clustered on geometry alone into eight regions (k-means on the coordinates, see Figure 4). Fold $k$ withholds all the counters of region $k$, which tests the extrapolation to a district without any counter. The same fold also withholds a scattered share of the counters of the other seven regions, which tests the interpolation between nearby sensors in a district the model has seen. Over the eight folds, every intra-muros representative counter (663) is a test counter once in extrapolation, and 346 of them are also withheld once in interpolation. The two regimes are reported separately throughout the paper to avoid a pooled score which would mix them in proportion to their counters. Both are needed as deployment in Paris is mostly a matter of spatial interpolation while transferring the architecture to a less-instrumented city is mostly a matter of extrapolation.

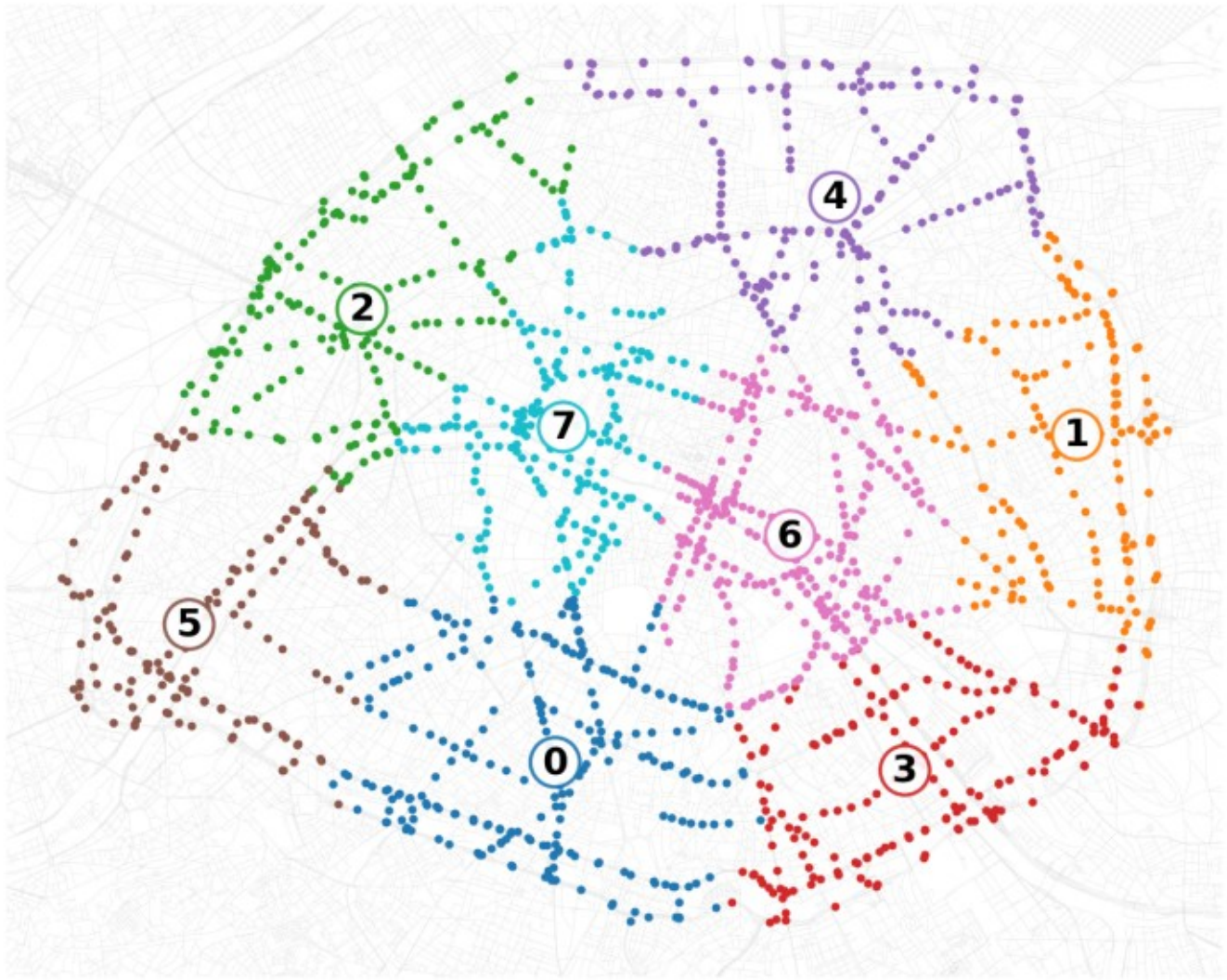

**Figure 4.** Illustration of the eight regions of the spatial cross-validation inside Paris intra-muros with the counting points colored by region (region $k$ is withheld as a whole in fold $k$).

Only measured and complete cells are used as targets and test values (not imputed, both directions counted), and the number of epochs is selected on the training counters only. The performance is reported as the hourly coefficient of determination $R^2$, the mean absolute error ($MAE$) and the bias at the held-out counters, as means over the eight folds with their standard deviation. Variants of the model are compared on the $MAE$ by paired tests over the eight folds (see Supplementary Section S8). In addition, the January–August 2026 period is kept outside the training record, so that the temporal generalization of the frozen model can be assessed separately (see Section 3.9).

### *2.8. Deployment*

For the deployment, the model is trained on all the sensors for the same number of epochs and produces the hourly flow of every segment for each year (8,759–8,783 h × 61,993 segments, in about 9 minutes per year on the GPU). The estimates are capped by the maximum flow observed on the instrumented segments of the same road class, which bounds the rare extrapolation excesses. In the final maps, the measured flows replace the estimates on the instrumented segments.

## 3. Results

### *3.1. Overall performance in interpolation and extrapolation*

As a first step of analysis, the overall held-out results on the eight folds are presented in Figure 5, as means over the folds with their standard deviation. In interpolation, the model reaches a held-out hourly $R^2$ of $0.89 \pm 0.07$ and an $MAE$ of $297 \pm 39$ veh h $^{-1}$, which corresponds to a quarter of the mean observed flow. In extrapolation, $R^2$ is $0.74 \pm 0.24$ and the MAE $355 \pm 88$ veh h $^{-1}$, a third of the mean flow. Over the pooled held-out predictions, four hourly values out of five are within a factor of two of the measurements in interpolation ($FAC2$ of 0.80) and three out of four in extrapolation (0.75). Fold means are reported rather than pooled values for $R^2$, since the pooled $R^2$ (0.91 and 0.86) includes the contrast between districts, which the model does not have to learn.

According to Figure 5C and D, the error is that of a spatial interpolator pulled towards its prior, with the lowest flows overestimated and the highest underestimated. The mean bias is -1% of the mean observed flow in interpolation and -6% in extrapolation. A model that uses these estimates will therefore inherit this slight underestimation, for instance in the evaluation of pollutant emissions such as $NO_x$.

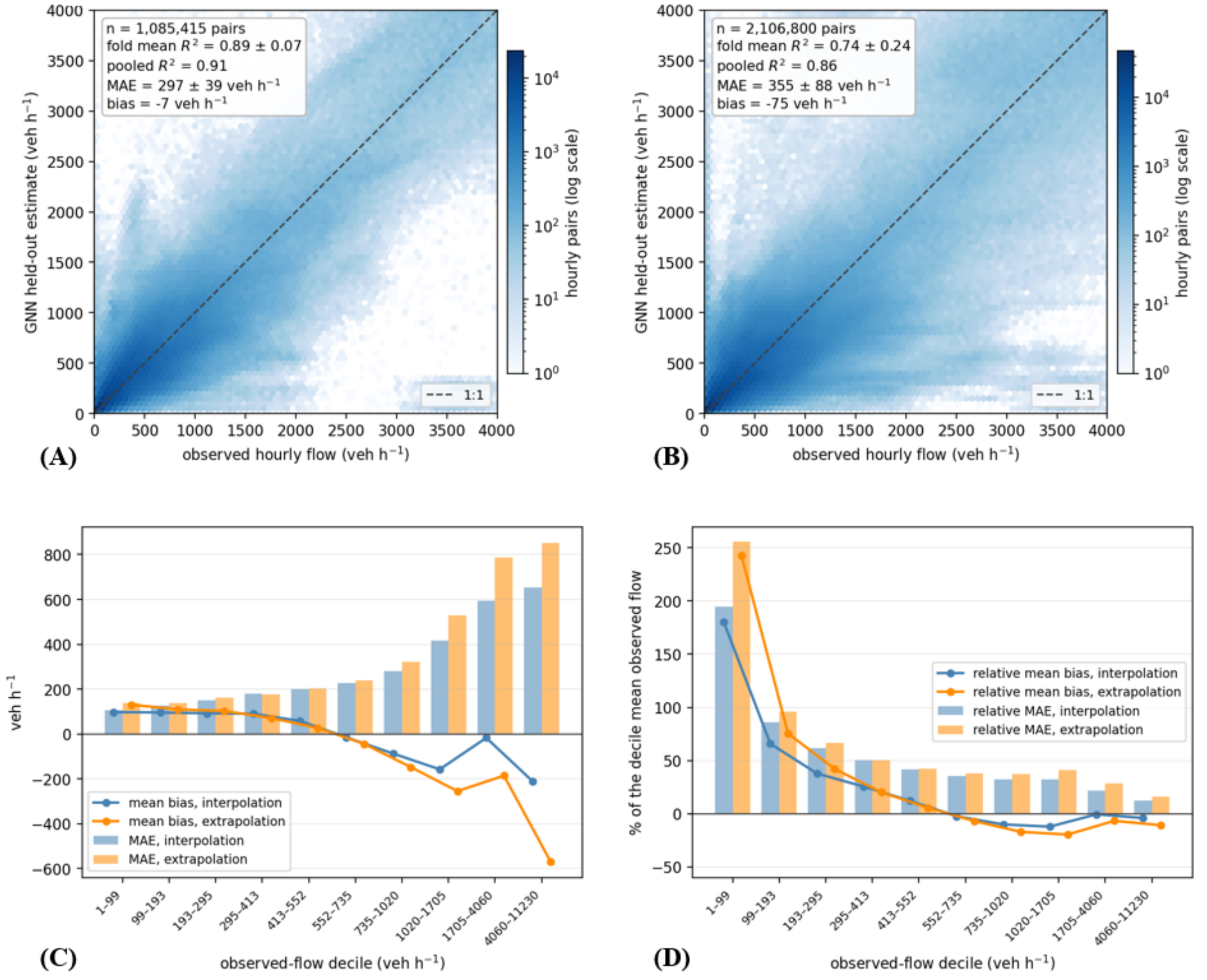

**Figure 5**. Traffic reconstruction at the held-out counters in Paris intra-muros, with the eight folds combined. Observed versus estimated hourly flows in (A) interpolation and (B) extrapolation (colors show the point density on logarithmic scale, and the dashed line is the 1:1 line). (C) Mean bias and mean absolute error ($MAE$) by decile of observed flow. (D) The same quantities expressed relative to the mean observed flow of each decile.

### *3.2. Variability between folds*

The results were then analyzed fold by fold. Since the withheld counters carry very different traffic from one fold to the next (mean observed flow from 580 to 1,670 veh/h), the folds are compared on the error relative to the mean observed flow of their test counters (Figure 6A). In interpolation the relative error is about a fifth to a third of the flow in seven folds out of eight, and reaches 42% in the fold whose test counters carry the lowest traffic. In extrapolation the spread is wider, and the fold-mean $R^2$ of 0.74 mixes six regions at or above 0.75 with the two central regions (folds 6 and 7) below 0.4 (Figure 6B). Two mechanisms explain this spread.

The first one concerns the regions that contain part of the ring road and of the urban motorways. The motorway sections carry flows about seven times those of ordinary streets, so that their absolute error, three to four times larger, dominates the $MAE$ and the bias of the fold. Relative to their own flow, however, they are estimated better than ordinary streets, with a relative error of about 20% against 45–50% in extrapolation. When the ring-road section of a region is withheld, its level has to be guessed from the sections of the other regions, and a section with higher or lower traffic than those is missed by that difference. In the south-eastern region (fold 3), the motorway counters are underestimated by about 870 veh/h, a third of their flow, which gives this fold the largest absolute error of all.

The second one concerns the two central regions, which contain no motorway counter. Their absolute error is below average but applies to the lowest mean flows, so that their relative error is the largest. Their $R^2$ collapses because the observed hourly flows of these regions vary little from one street to another (standard deviation of about 500 veh/h against 1,200–1,900 veh/h elsewhere, see the $\sigma$ labels of Figure 6B). Part of the relative error of ordinary streets reflects the natural variability of low flows rather than the reconstruction. Divided by the standard deviation of each counter, the held-out error is about 0.7 for flows between 150 and 2,500 veh/h in both regimes, and 0.3–0.4 on the busiest axes. Only the few counters below 150 veh/h exceed two standard deviations, so the overestimation of the quietest streets is a genuine error of the reconstruction (see Supplementary Table S6). The model thus extrapolates least well, in relative terms, to the dense centre of ordinary streets and, in absolute terms, to the regions whose motorway sections differ most from the rest of the city.

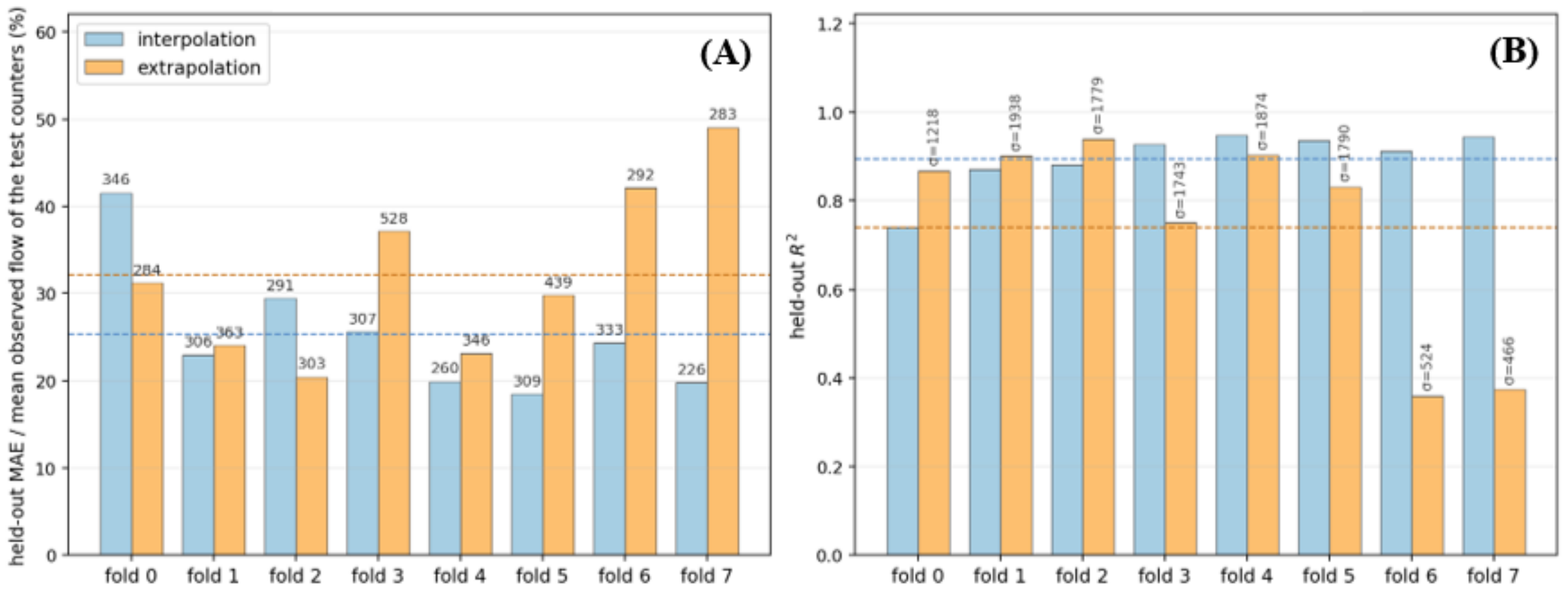

**Figure 6.** Results at the held-out counters for each fold in Paris intra-muros (regions shown in Figure 4). (A) $MAE$ for each fold, relative to the mean observed flow at the held-out counters of that fold (labels give the absolute $MAE$ in veh/h). (B) Held-out $R^2$ for each fold. ($\sigma$ denotes the standard deviation of the observed hourly flow at the held-out counters in veh/h). Dashed lines show the mean over the eight folds.

### *3.3. Contribution of the model components*

In order to assess the contribution of each component of the model, the climatology alone, the network trained without the offset and the full model were compared on the same folds, targets and number of training epochs. The corresponding results are gathered in Table 1. These runs, like those of Section 3.4 and 3.5, use three diffusion layers, a configuration equivalent to the deployed two-layer model (see Section 3.6). According to these results, the climatology brings nothing in interpolation, where the network alone is as good as the full model (paired $p$=0.77). In extrapolation, both components are middling alone, and their combination tends to be better than either of them (removing the climatology adds 27 veh/h on average and degrades six folds out of eight: a difference that is not significant, $p$=0.18). The offset can therefore be seen as an insurance for the regime where the network has no neighbour to rely on rather than as a precision gain, and is kept for deployment.

**Table 1.** Errors at the held-out counters for the traffic model and for each of its two components in Paris intra-muros, in interpolation and extrapolation. Values are the $MAE$ (veh/h) and $R^2$, given as mean ± standard deviation over the eight folds. All rows use the same folds, targets and number of training epochs.

| Model | Interpolation | | Extrapolation | |
|---|---|---|---|---|
| | $MAE$ | $R^2$ | $MAE$ | $R^2$ |
| Climatology only | $401 \pm 46$ | $0.81 \pm 0.08$ | $415 \pm 77$ | $0.61 \pm 0.38$ |
| GNN without climatology | $291 \pm 28$ | $0.91 \pm 0.05$ | $395 \pm 108$ | $0.68 \pm 0.27$ |
| Full model, three layers | $296 \pm 41$ | $0.89 \pm 0.08$ | $368 \pm 90$ | $0.73 \pm 0.24$ |
| **Full model, two layers (deployed)** | $297 \pm 39$ | $0.89 \pm 0.07$ | $355 \pm 88$ | $0.74 \pm 0.24$ |

### *3.4. Importance of the inputs*

The inputs that matter were then identified by permutation (see Figure 7 and Supplementary Section S3). To do so, one block of input columns is shuffled, and the same trained model is scored again. A block whose shuffling changes nothing is unused, whereas a block whose shuffling degrades the results is used, although a model retrained without it might compensate. Based on these results, the road attributes (class, lanes, width, speed, length, one-way, urban, bridge, tunnel, sinuosity) carry almost all of the spatial information. The eighteen other static features matter a little in interpolation and not at all in extrapolation, since the local context sharpens the estimate next to a known counter but does not transfer to an unseen district. Besides the road atrtibutes, what transfers is temporal, as the time cycles and the live state of the counting network are the two main temporal inputs in both regimes. The seven weather features add less than 1 veh/h, which is not surprising since the live state of the network already contains, observed rather than predicted, any effect of the weather on traffic: they would presumably regain some value in a city without a live counting network. Roadworks, strikes and the daylight flag have no visible effect on the average error either:

roadworks and strikes are rare or local events whose effect, if any, is confined to the hours and segments concerned, and daylight is redundant with the time cycle.

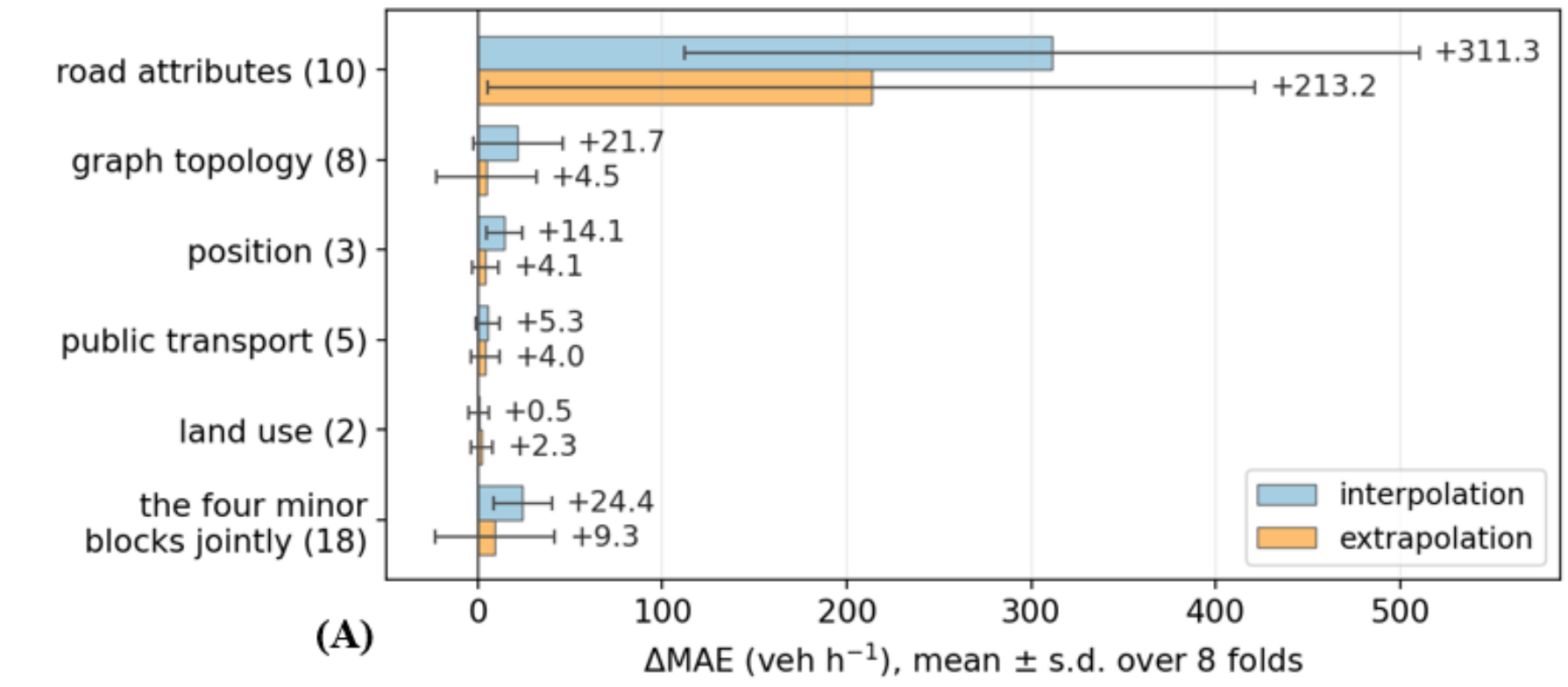

(A)

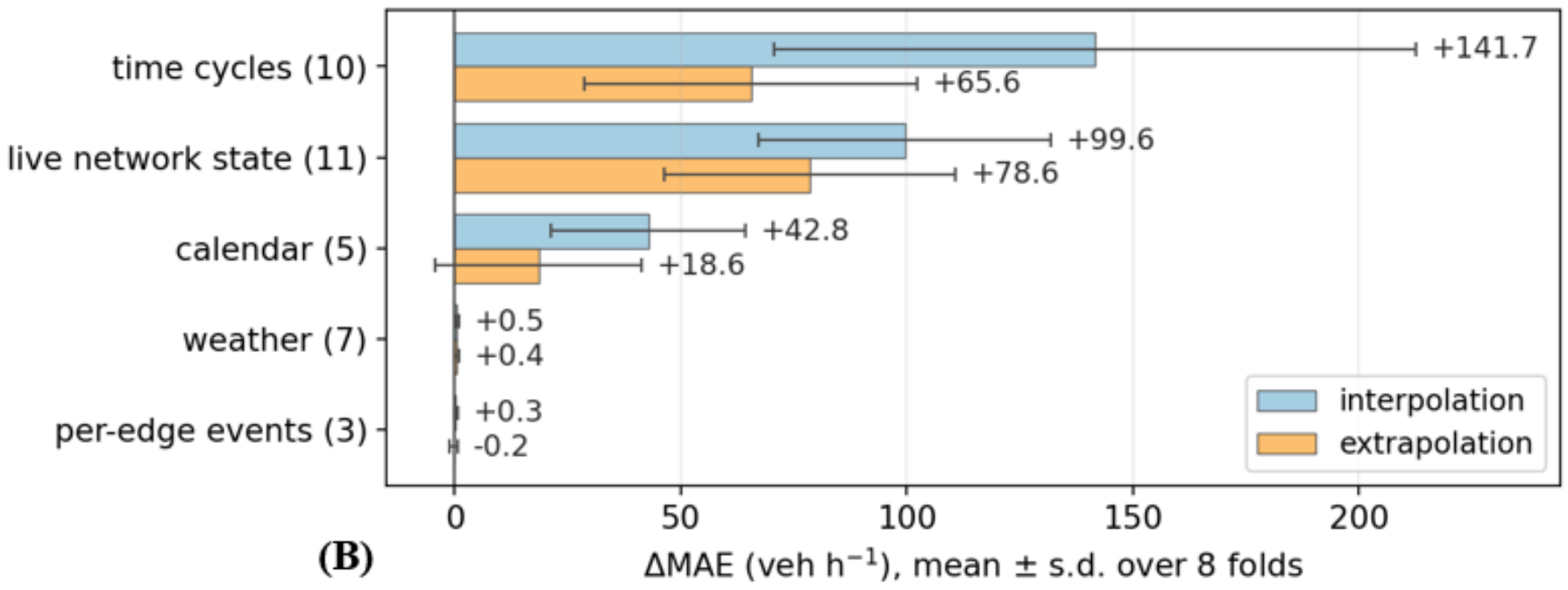

(B)

**Figure 7**. Permutation importance of the input features of the traffic model, computed with all counters and a model with three diffusion layers. The panels show the increase in held-out $MAE$ when one block of (A) spatial or (B) temporal features is randomly shuffled, without retraining the model. Values are mean ± standard deviation over the eight folds. The number of features in each block is given in brackets. The climatology is not shuffled.

### *3.5. Masking strategy*

The way the counters are hidden during training was also compared on the eight folds, at equal numbers of masked counters, considering scattered counters, whole regions, or a mix of both. The corresponding results are given in Table 2. Region masks degrade interpolation (+17 veh/h, 7 folds out of 8, $p$=0.04) and improve extrapolation by 3–7 veh/h, an improvement that the eight folds are not sufficient to establish. Since most of the unmonitored intra-muros network lies within a few hops of a counter (see Section 2.3), deployment in Paris is essentially a spatial interpolation problem and scattered masks were kept. The sign of the effect in extrapolation nevertheless suggests that region masks would be preferable for a transfer to a less-instrumented city.

**Table 2.** Effect of the masking strategy used during training on the held-out error of the traffic model (Paris intra-muros, eight folds, three diffusion layers). Both strategies mask the same number of counters and use the same climatology. Values are the difference in $MAE$ (veh/h) between region masks and scattered random masks. Positive values mean that region masks increase the error. "Folds" gives the number of folds, out of eight, in which region masks perform worse. $p$: p-value of a paired $t$-test over the folds (Wilcoxon test in brackets).

| Regime | Interpolation | | | Extrapolation | | |
|---|---|---|---|---|---|---|
| | $\Delta MAE$ | Folds | $p$ | $\Delta MAE$ | Folds | $p$ |
| mixed (0.5) | +11.8 | 5/8 | 0.141 (0.195) | -7.1 | 3/8 | 0.32 (0.31) |
| regions only (1) | +16.9 | 7/8 | 0.042 (0.078) | -3.2 | 4/8 | 0.73 (0.95) |

### *3.6. Hyper-parameters inherited from the model family*

Three settings were inherited from GNNUI rather than chosen for Paris, namely the depth of the diffusion encoder (three layers in GNNUI), the Chebyshev order ($K$=3) and the masking rate (25%). Each of them was varied in turn on the eight folds and compared to the deployed configuration (two layers, $K$=3, 25%) by paired tests. The results are given in Table 3.

Firstly, the depth of the encoder has no significant effect. From one to four layers, that is a reach of 3 to 12 hops and 2.4 times more parameters, the held-out error is flat and no paired difference is significant. Even the one-layer network, whose three-hop reach leaves half of the unmonitored network out of range, loses little, and its small deficit (10-20 veh/h) is spread evenly over the distance bins (see Table 4 in Section 3.7). The information that the graph brings to a segment is therefore local. The counters adjacent to its junctions and the segments within a few hops say most of what the network has to say about it, and a reach of twelve hops does not estimate the segments beyond five hops of any counter any better than a reach of six. Deeper encoders only fit the training data better. The deployed model has two layers because, once the per-fold error is weighted by the hop-distance distribution of the unmonitored network, two layers beat three and four in extrapolation in seven folds out of eight and are equivalent in interpolation.

**Table 3**. Sensitivity of the traffic model to its hyper-parameters (Paris intra-muros, eight folds). Each variant changes a single setting of the deployed configuration. Values are the held-out MAE (veh $h^{-1}$, mean ± standard deviation over the folds) and the paired difference with the deployed configuration. This difference is given with the number of folds in which the variant performs worse and the $p$-value of a paired $t$-test.

| Setting | Variant | Interpolation | | Extrapolation | |
|---|---|---|---|---|---|
| | | $MAE$ | Δ (folds, $p$) | $MAE$ | Δ (folds, $p$) |
| | **deployed**: 2 layers, $K$=3, 25% | $297 \pm 39$ | - | $355 \pm 88$ | - |
| Depth | 1 layer (reach 3 hops) | $306 \pm 46$ | +9 (6/8, 0.30) | $374 \pm 108$ | +19 (5/8, 0.18) |
| | 3 layers (9 hops) | $296 \pm 41$ | -1 (2/8, 0.89) | $368 \pm 90$ | +13 (5/8, 0.11) |
| | 4 layers (12 hops) | $298 \pm 39$ | +0 (3/8, 0.95) | $360 \pm 97$ | +5 (4/8, 0.53) |
| Order | $K$=2 (2 layers, 4 hops) | $304 \pm 41$ | +7 (6/8, 0.44) | $360 \pm 85$ | +6 (5/8, 0.57) |
| | $K$=4 (2 layers, 8 hops) | $308 \pm 55$ | +11 (5/8, 0.25) | $364 \pm 93$ | +9 (5/8, 0.29) |
| | $K$=2 (3 layers, 6 hops) | $296 \pm 30$ | -1 (5/8, 0.85) | $365 \pm 96$ | +10 (5/8, 0.29) |
| Masking rate | 15% | $303 \pm 43$ | +6 (5/8, 0.14) | $379 \pm 104$ | +24 (6/8, 0.020) |
| | 40% | $293 \pm 37$ | -5 (4/8, 0.43) | $361 \pm 82$ | +6 (4/8, 0.28) |

Secondly, the Chebyshev order was kept at $K$=3, since at two layers it gives the lowest mean $MAE$ in both regimes, and extending the reach at fixed depth changes the error by a few veh/h only, without any significant difference.

Lastly, the masking rate is the only setting with a clear effect, and this effect is confined to extrapolation. Hiding 15% of the counters instead of 25% costs +24 veh/h ($p$=0.02, and +26 in eight folds out of eight on the deployment-weighted error, $p$=0.0004), and hiding 40% costs +6 (+16 on the deployment-weighted error, $p$=0.044). With too few counters hidden the task is too easy and the network does not learn to do without its instrumented neighbours, whereas with too many the context it needs is destroyed. The 25% selected for the deployed model is thus a compromise, and it is the first setting to revisit for a transfer to a less-instrumented city.

### *3.7. Performance against the distance to the counters*

Since most of the deployment is near-field (see Section 2.3), the held-out error was also analyzed as a function of the graph distance to the nearest training counter. The corresponding results are presented in Figure 8. According to this figure, the error is flat up to five hops, then increases by about a fifth (the bias, not shown, turns negative), as the model falls back towards its climatology. This step is not a limitation of the receptive field, since a four-layer network with a reach of twelve hops shows the same step, and a one-layer network with a reach of three hops shows a deficit of 10–20 veh/h in every bin, including the bins within its reach (see Table 4). The same shape is observed in metric distance (see Figure 10A), with a far field beyond 2 km that concerns only a small share of the network.

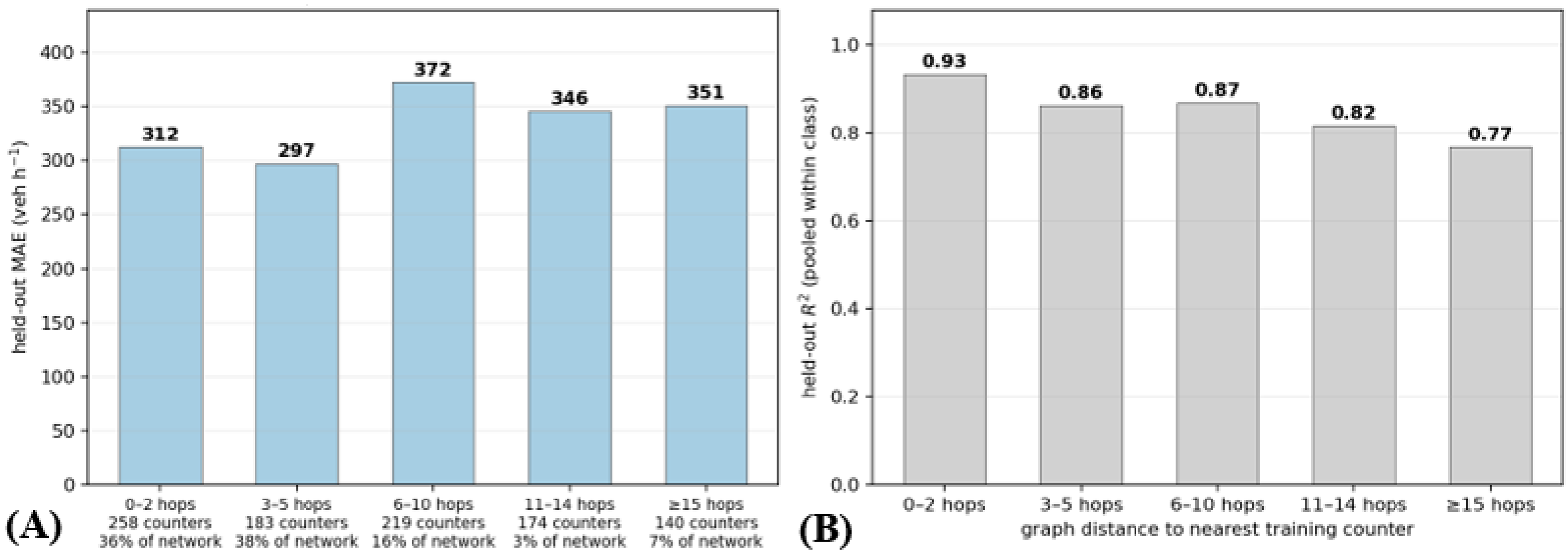

**Figure 8**. Held-out performance as a function of the graph distance (number of hops) to the nearest training counter, with the eight folds combined (Paris intra-muros). (A) $MAE$ for each distance class. (B) Held-out $R^2$ computed on the pooled data of each class.

**Table 4**. Held-out $MAE$ (veh/h) as a function of the graph distance to the nearest training counter, for four encoder depths (Paris intra-muros, eight folds combined). The reach of the encoder, in hops, is three times its depth.

| Depth (reach) | 0–2 hops | 3–5 | 6–10 | 11–14 | ≥15 |
|---|---|---|---|---|---|
| 1 layer (3) | 323.3 | 317.4 | 393.4 | 355.7 | 361.5 |
| **2 layers (6, deployed)** | **312.4** | **297.3** | **372.4** | **345.6** | **351.0** |
| 3 layers (9) | 314.8 | 298.7 | 397.1 | 350.7 | 352.2 |
| 4 layers (12) | 319.4 | 299.3 | 373.9 | 351.5 | 350.8 |
| 1 layer - 2 layers | +10.9 | +20.1 | +21.0 | +10.1 | +10.5 |

### *3.8. Reconstructed traffic over the network*

Figure 9 shows the result of the deployment for 2025, with, on the one hand, the counted segments on the main axes and, on the other hand, the hourly flow reconstructed on every other segment of the city.

The mean estimated flow on the unmonitored segments spans the whole road hierarchy (see Figure 10C and Supplementary Table S5), and in five classes out of six the unmonitored mean is below the instrumented mean, which is expected since the counters are placed on the busier axes of each class. Local streets have only 59 counters for 15,698 unmonitored segments and still receive plausible magnitudes, correctly ordered relative to the classes above them. Within the day (Figure 10B), the model reproduces the daytime cycle and the depth of the trough at 04:00 on the unmonitored edges, but keeps slightly too much traffic in the evening and the first part of the night. The night share (22:00–05:00) of the unmonitored traffic exceeds that of the measured network by 8% in relative terms, which is a regression towards the climatology at hours the sensors constrain weakly. This is the only systematic shape error of the reconstruction.

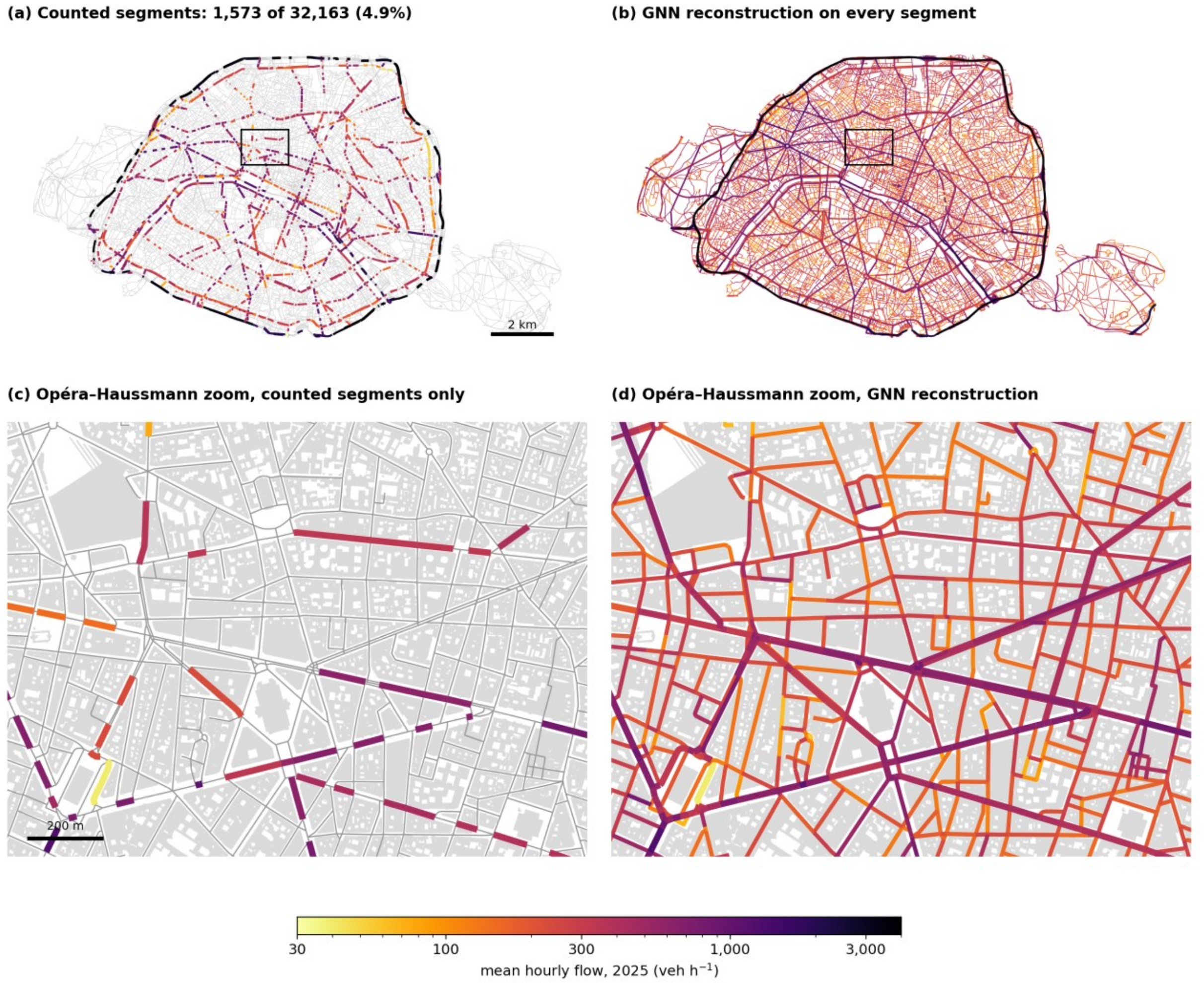

**Figure 9**. Traffic flows in Paris intra-muros in 2025. (a) Mean measured flow on the segments equipped with a counter, shown over the road network in grey. (b) Mean hourly flow on every segment of the final maps, measured on the segments with a counter and estimated elsewhere. (c,d) The same views for the 1.6 × 1.2 km Opéra–Haussmann area outlined in (a) and (b), with buildings in light grey.

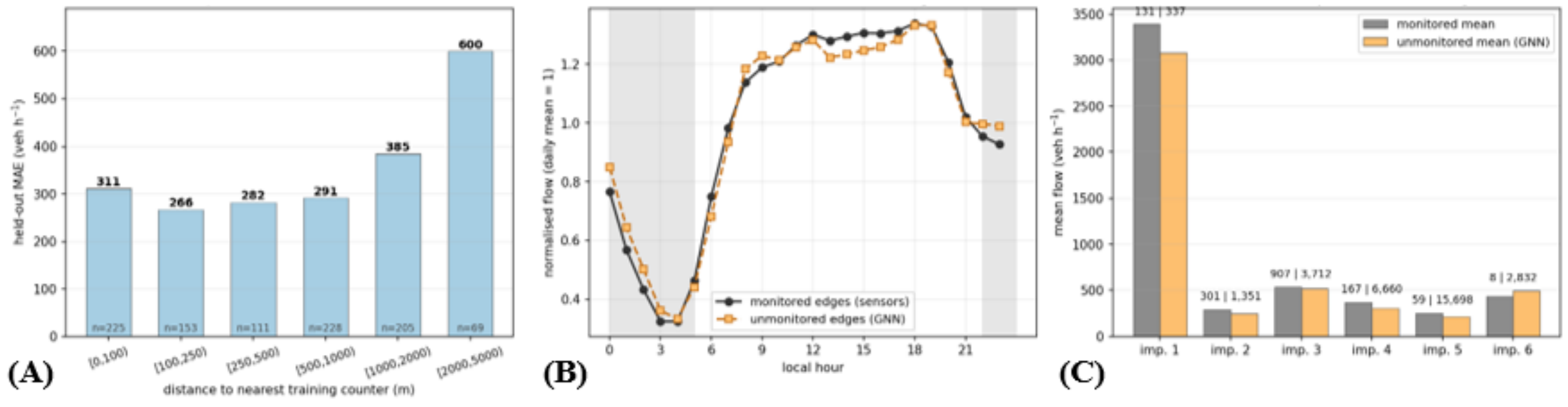

**Figure 10**. Diagnostics of the traffic reconstruction in Paris intra-muros. (A) Held-out $MAE$ as a function of the distance to the nearest training counter, with the eight folds combined ([n]: number of held-out counters). (B) Normalised daily flow

profiles of the segments with and without a counter in the 2025 maps (the shaded area covers 22:00–05:00). (C) Mean flow in 2025 by BD TOPO importance class, for segments with and without a counter.

### *3.9. Evaluation outside the training period*

As a final analysis, the deployed model, frozen and fed with the 2026 counter data, was scored month by month by masked reconstruction, as in training. The counters are split into four random groups of 25%, each group is hidden in turn and estimated, so that every counter is scored at every hour with its own measurement invisible, and this in every year alike. This in-sample score (MAE of about 100 veh/h, against 300 to 350 veh/h at held-out counters) is not comparable to the held-out performance of the spatial protocol, but it isolates the temporal transfer. According to Figure 11, on January–August 2026, the relative error of the frozen model stays within the band of the same months of 2022–2025, except in the first two months where it is just above it. It should nonetheless be noted that only 1,376 of the 1,579 counters still report in 2026, and the imputation flags of the provider are not available for 2026, so that the score and its band are computed without the imputed-cell criterion.

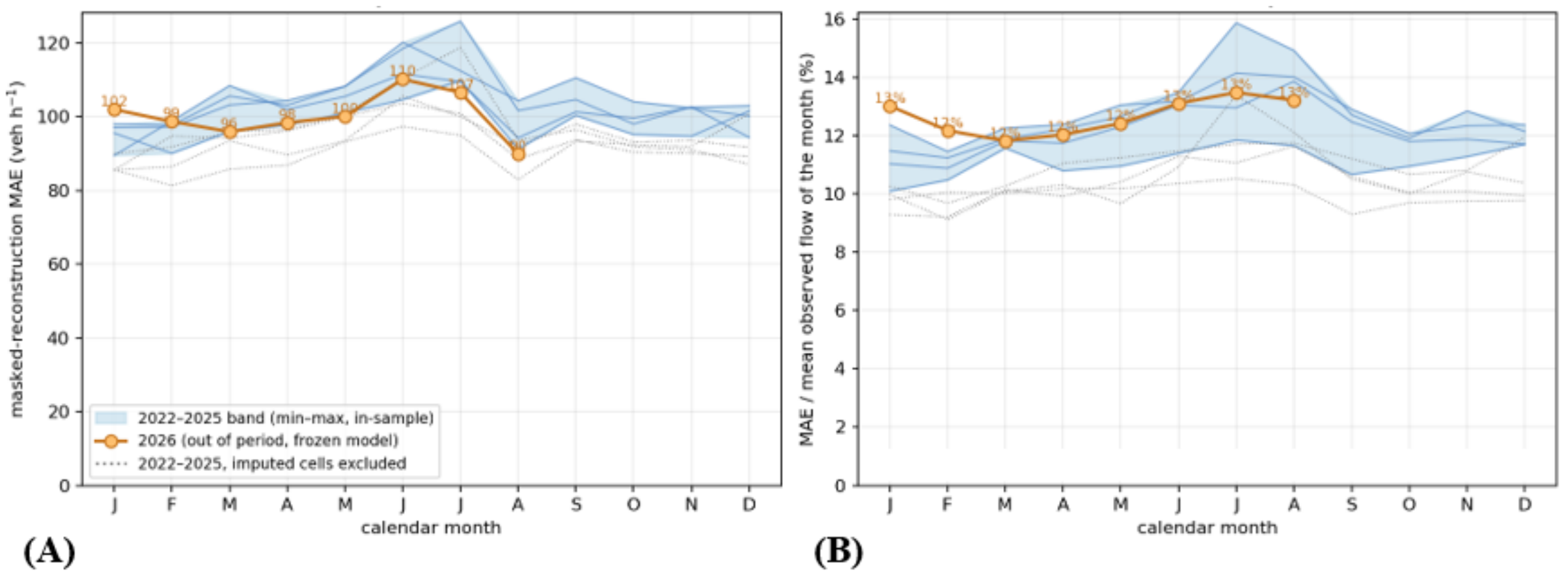

**Figure 11**. Monthly reconstruction error of the trained traffic model at the masked counters, without retraining (Paris intra-muros). (A) $MAE$ and (B) MAE relative to the mean observed flow of the month. For 2022–2025, the shaded band shows the range between the minimum and maximum of each calendar month, and the thin lines show the individual years. The dotted lines show the same years with the imputed cells excluded. The orange line shows January–August 2026.

## 4. Discussion

Two choices of the evaluation protocol determine what the scores mean. The counters that share a published series were merged before the split, because a hold-out that separates them places the twin of a test counter in the training set and inflates the scores. In addition, interpolation and extrapolation were scored separately, with fold means, because a pooled score mixes the two regimes in proportion to their counters and is inflated by the contrast between districts. Both precautions are recommended for any spatial cross-validation on an urban counting network.

Cross-study comparisons are indicative only, since the datasets, the targets and the protocols differ. The state of the art for unmonitored local roads under spatially informed cross-validation reports $R^2$ ranging from 0.67 to 0.79 for annual average daily traffic (Ma et al., 2026; Sfyridis & Agnolucci, 2020), and the inductive kriging literature evaluates on sensor or trajectory benchmarks with near-complete coverage (Kaiser et al., 2026; Wu, Zhuang, Lei, et al., 2021; Zheng et al., 2023). The present model reaches R2 values in the same range for hourly flows on a real network with only 4.9% coverage, with the two regimes separated and open data only. These values are, however, driven by the main axes and by the daily cycle, and are not directly comparable with annual scores on local roads.

The approach has, however, some limitations which can be summarized as follows.

- The validation covers the roads that the counters cover, which are overall the busiest ones. In extrapolation, 115 counters are on the motorways and the ring road (class 1), 467 on the axes (classes 2 and 3), 57 on the local distributors (class 4) and 19 on the local streets (class 5), which make up 11,411 and 33,566 segments of the graph respectively, more than two thirds of it (see Supplementary Table S7). The relative error follows the flow level more than the class. It is 18% on the motorways, 37% on the secondary axes and the local distributors, and 46–51% on the two classes whose counters carry less than 300 veh/h, the class-2 axes and the local streets, where the flows are also underestimated by 14–22%. The local distributors are thus reconstructed with the same relative error as the secondary axes (37%) but with a positive bias +14%), which is encouraging for the unmonitored network, but the local streets, which are most of it, are verified on 19 sites only, and these sites are where the reconstruction is weakest.

- The error is that of an interpolator pulled towards its prior, with a negative bias where the model extrapolates. The local streets are constrained by the road attributes and the climatology only, and the climatology of these classes inherits the sitting bias of the counters.
- The out-of-period test is an in-sample masked reconstruction and not a spatial hold-out.

Some improvements could therefore be considered. The masking rate is the first setting to revisit for a transfer to a less-instrumented city, and region masks the second. The night-time excess of the unmonitored edges (see Section 3.8) could be reduced by weighting the night hours in the training objective or by rescaling their diurnal profile on that of the measured network. Two kinds of data would give the local streets the observation that the counters do not provide. Floating car data, the speeds and trajectories of GPS-equipped fleets, cover the small streets precisely where the counters are absent (Herrera et al., 2010), and an origin–destination matrix would give the model a demand prior in place of the climatology (Ortúzar & Willumsen, 2011). Neither was considered in this study, which was limited to open data, but both could enter the model naturally as additional inputs of the segments. Lastly, further research is required to apply the approach to other cities with different counting networks, in order to assess its generalization beyond the Paris case, and in particular to cities where the coverage is lower and the extrapolation regime dominates.

## 5. Conclusion

In this study, a graph neural network was adapted to the sparse counting network of Paris in order to reconstruct hourly traffic flows on every segment of the city, and evaluated with a spatial cross-validation designed for a real counting network. The main conclusions are as follows.

(a) The model reconstructs the hourly flows with a held-out error of a quarter of the mean flow where it interpolates between nearby counters and a third where it extrapolates to a whole district without any counter, using open data only.

(b) The climatology offset mainly acts as an insurance for the districts without counters, while the road attributes, the time cycles and the live state of the counting network carry most of the performance: the weather inputs have a negligible effect.

(c) The depth of the diffusion encoder has no significant effect, whereas the masking rate is the only inherited setting with a clear effect.

(d) The frozen model remains within or just above the error band of the previous years on a period outside its training record, for the reconstruction of masked training counters.

# Supplementary material

## S1. Input features of the traffic model

The deployed model takes four feature families per hourly snapshot, namely a 28-dimensional static block per edge, 7 features per node (3 static and 4 observed-context), 4 dynamic features per edge, and 36 temporal features (33 global and 3 per edge). All continuous features are z-scored; missing static attributes are filled with class defaults (rank 5, 1 lane, 7 m width, 50 km $h^{-1}$) before scaling.

**Table S1.** Input features of the traffic model, by family.

| Family | *n* | Features | Source |
|---|---|---|---|
| Edge, static | 28 | road rank, lanes, carriageway width, regulatory speed, length, one-way, urban, bridge, tunnel, sinuosity; betweenness centrality, mean neighbour centrality, graph-bridge flag, class-transition flag, mean and max neighbour rank within 2 hops, local road density, distance to centre, distance to nearest importance-1/2 road, bearing (sin, cos); land use, population density; proximity to heavy transit and to bus stops, bus stops within 50 m, transit stops within 200 m, tram intersection | IGN BD TOPO v3.5 and derived graph; OSM; INSEE; IDFM |
| Node, per hour | 7 | normalised easting and northing, degree; mean and max normalised observed flow on adjacent unmasked sensor edges, number of adjacent active sensors, masked-neighbour flag | BD TOPO; counting network |
| Edge, dynamic | 4 | normalised observed flow and loop occupancy (unmasked sensor edges, else 0), mask flag, missing flag | Counting network |
| Temporal, global | 33 | hour, day of week, day of month, month, week of year (sin/cos each); weekend, public holiday, holiday eve, school holiday, bridge day; temperature and precipitation with their 1 h and 3 h lags, wind speed; live network mean, 1-week and 4-week rolling means; city-flow lags 1–6 h and trends 1 h, 3 h | Timestamp; holidays; Meteostat (Montsouris); counting network |
| Temporal, per edge | 3 | active roadworks (10 m buffer; 88,312 polygons 2022–2025), transit-strike day (46 dates), daylight | opendata.paris.fr; curated list; astral |

## S2. Departures from GNNUI

Table S2 lists the main differences between GNNUI from Kaiser et al. (2026) and the present model.

**Table S2.** Main differences between GNNUI and the present model.

| Aspect | GNNUI | This work |
|---|---|---|
| Graph | Segments as nodes; adjacency between segments that meet | BD TOPO junction graph: junctions as nodes (44,763), segments as edges (61,993), ordered for decoding; two Chebyshev random walks on the undirected graph, one row-normalised, the other column-normalised |
| Prediction head | Output layer on the node embedding | Edge decoder on the concatenation of the two FiLM-conditioned endpoint embeddings, the 28 static and 4 dynamic descriptors (encoded to 32) and the 36 temporal features (352 values); MLP 352 → 128 → 64 with skip connection and LayerNorm |
| Temporal conditioning | Temporal features appended to node inputs | FiLM rescaling of both endpoint embeddings by a 36-feature encoding (33 city-wide, 3 per edge) |

| Aspect | GNNUI | This work |
|---|---|---|
| Output scale | ZINB parameters learned from scratch | ZINB with a climatology log-offset on $\log n$ (median per road rank × hour × weekend, ramps apart, training sensors only) |
| Masking | Random subset of segments | 25% of sensors per step, scattered (deployed); region masks (k-means blobs of the training sensors) evaluated as an alternative (Table 2 of the main text) |
| Extra supervision | None (full-coverage benchmarks) | +371 virtual sensors by label propagation along junction-free chains, never evaluated |
| Validation | 10% spatial hold-out | 8-fold spatial CV on 790 groups of counters sharing a series; each fold scores interpolation (withheld counter, seen region) and extrapolation (withheld region) separately; measured and complete cells only; paired tests by fold |

## S3. Departures from GNNUI

One block of input columns is shuffled across edges (spatial blocks) or across hours (temporal blocks) and the unchanged checkpoint is scored again. The climatology is not shuffled. A null effect proves that a block is unused. A strong effect does not prove that removing it would degrade, since blocks can be redundant. The screening is computed on the whole counter set and the eight folds, with the three-layer configuration (equivalent to the deployed two-layer model, Section 3.6 of the main text).

**Table S3.** Permutation screening of the input features of the traffic model (three-layer configuration). Increase of the held-out MAE when one block is shuffled, in interpolation and extrapolation.

| | Block ($n$ features) | Interpolation | | | Extrapolation | | |
|---|---|---|---|---|---|---|---|
| | | $\Delta$MAE | Folds | $p$ | $\Delta$MAE | Folds | $p$ |
| Spatial | road attributes (10) | $+311 \pm 199$ | 8/8 | 0.003 | $+213 \pm 208$ | 8/8 | 0.023 |
| | graph topology (8) | $+22 \pm 24$ | 6/8 | 0.040 | $+4 \pm 27$ | 5/8 | 0.65 |
| | position (3) | $+14 \pm 10$ | 7/8 | 0.005 | $+4 \pm 7$ | 4/8 | 0.15 |
| | public transport (5) | $+5 \pm 7$ | 7/8 | 0.054 | $+4 \pm 8$ | 5/8 | 0.20 |
| | land use (2) | $+1 \pm 5$ | 4/8 | 0.79 | $+2 \pm 6$ | 4/8 | 0.31 |
| | the four minor blocks jointly (18) | $+24 \pm 16$ | 8/8 | 0.003 | $+9 \pm 32$ | 4/8 | 0.44 |
| Temporal | time cycles (10) | $+142 \pm 71$ | 8/8 | 0.0008 | $+66 \pm 37$ | 8/8 | 0.0015 |
| | live network state (11) | $+100 \pm 32$ | 8/8 | 0.0001 | $+79 \pm 32$ | 8/8 | 0.0002 |
| | calendar (5) | $+43 \pm 21$ | 8/8 | 0.0008 | $+19 \pm 23$ | 6/8 | 0.055 |
| | weather (7) | $+0.5 \pm 0.4$ | 7/8 | 0.015 | $+0.4 \pm 0.5$ | 5/8 | 0.085 |
| | per-edge events (3) | $+0.3 \pm 0.4$ | 6/8 | 0.13 | $-0.2 \pm 0.9$ | 4/8 | 0.52 |

## S4. Training and inference settings

**Table S4.** Settings of the traffic model.

| Setting | Value |
|---|---|
| Parameters | 367,459 (two diffusion layers) |
| Node encoder | MLP 7 → 128 → 128; two diffusion convolutions (Chebyshev $K = 3$, forward and backward walks), residual and LayerNorm |
| Edge decoder | Temporal encoder 36 → 64 → 32; FiLM per endpoint ($\gamma, \beta$ from 32 → 64 → 2 × 128); concatenation (352) → MLP 128 → 64 with skip connection (352 → 64) and LayerNorm; ZINB |

| Setting | Value |
|---|---|
| | head (64 → 3) |
| Training | AdamW (Loshchilov & Hutter, 2019), learning rate $5 \times 10^{-4}$, cosine schedule, 8 epochs; 25% of the sensors masked per step; seed 42 |
| Valid targets | Measured cells only (not imputed by the provider), both directions counted |
| Deployment | Retrained on all sensors; estimates capped at the maximum observed on the instrumented segments of the same road class |

## S5. Network composition and flow by class

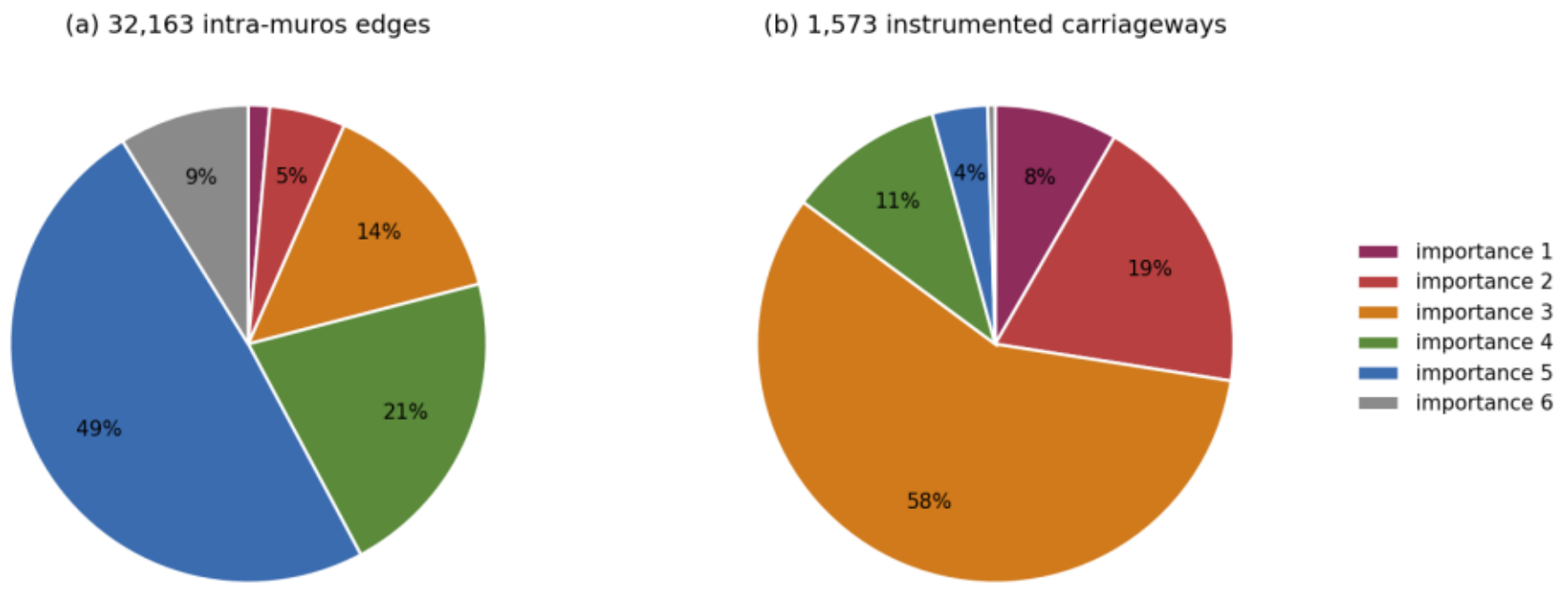

**Figure S1.** Composition of the Paris intra-muros road graph (a) and of its counting network (b) by BD TOPO importance class.

**Table S5.** Annual mean hourly flow by road class (BD TOPO importance, 1 = motorway class to 6 = lowest, heterogeneous class), Paris intra-muros, 2025 deployment: instrumented segments (measured) vs unmonitored segments (GNN estimate, mean and median over edges of the annual mean).

| Importance | *n* edges | *n* sensors | coverage | measured mean | GNN mean | GNN median |
|---|---|---|---|---|---|---|
| | | | | (veh/h) | (veh/h) | (veh/h) |
| 1 | 468 | 131 | 28.0% | 3,391 | 3,081 | 3,519 |
| 2 | 1,652 | 301 | 18.2% | 293 | 247 | 225 |
| 3 | 4,619 | 907 | 19.6% | 534 | 519 | 462 |
| 4 | 6,827 | 167 | 2.4% | 370 | 302 | 270 |
| 5 | 15,757 | 59 | 0.4% | 255 | 212 | 191 |
| 6 | 2,840 | 8 | 0.3% | 435 | 497 | 476 |
| all | 32,163 | 1,573 | 4.9% | – | – | – |

## S6. Diffusion convolution

Let $A$ be the adjacency matrix of the junction graph after symmetrisation (each segment connects its two end junctions in both directions) and $D$ its diagonal degree matrix, zero degrees being clamped to one. The two propagation operators are

$$A_f = D^{-1}A, \qquad A_b = AD^{-1} = A_f^{\top}.$$

For a node-feature matrix $H$, the aggregation $A_f H$ weights each contribution by the inverse degree of the receiving junction, whereas $A_b H$ weights it by the inverse degree of the contributing neighbour, i.e. two normalisations of the same undirected connectivity. At layer $\ell$, with $H^{(\ell)} \in \mathbb{R}^{N\times 128}$, the layer forms

$$Z^{(\ell)} = \left[H^{(\ell)} \parallel T_1(A_f)H^{(\ell)} \parallel \dots \parallel T_K(A_f)H^{(\ell)} \parallel T_1(A_b)H^{(\ell)} \parallel \dots \parallel T_K(A_b)H^{(\ell)}\right],$$

where $T_1(M) = M$ and $T_k(M) = 2MT_{k-1}(M) - T_{k-2}(M)$ with $T_0(M) = I$, so that the polynomial of order $k$ combines information along paths of at most $k$ hops. The update is

$$Q^{(\ell)} = \mathrm{ReLU}\left(\mathrm{Dropout}\left(Z^{(\ell)}\right)W^{(\ell)} + 1\left(a^{(\ell)}\right)^{\mathrm{T}}\right), \qquad H^{(\ell+1)} = H^{(\ell)} + \mathrm{LayerNorm}\left(Q^{(\ell)}\right),$$

with $W^{(\ell)} \in \mathbb{R}^{(2K+1)128\times 128}$ and a bias $a^{(\ell)}$; dropout (0.1) is active during training only. For $K = 3$ the concatenated matrix has 896 columns. Two such layers give a receptive field of six hops around each junction; the context available to a segment is the union of the receptive fields of its two end junctions. The three decoder outputs are mapped to valid parameters by $n = \exp(\cdot)$, $p = \sigma(\cdot)$ and $\pi = \sigma(\cdot)$, with numerical clipping of the arguments and results in the implementation.

### S7. Climatology construction

The climatological baseline groups the training observations by road rank, hour of day and weekday/weekend status. Depending on the fold, six to seven rank classes are represented, motorway ramps forming a class of their own, so that the lookup table contains $(6–7) \times 24 \times 2$, i.e. 288 to 336 cell medians. An empty cell takes the global training median, and the baseline is floored at 1 veh/h before its logarithm is taken. Each validation fold has its own table, estimated from the training sensors of that fold only, so that no held-out observation enters it; for deployment the table is recomputed from all the sensors. The offset is applied to the first decoder output, $\log(n) = \eta + \log(b)$, which also scales the size parameter of the negative binomial.

### S8. Statistical comparison of the model variants

Variants of the model (components, inputs, masking regime, hyper-parameters) are compared on the held-out $MAE$ by paired tests over the eight folds, each fold being its own control. Counters are never pooled across folds for these tests, since they are correlated within a fold. A paired $t$-test is run with a Wilcoxon signed-rank test alongside, and a Friedman test when more than two variants are compared at once; the tables of the main text give the mean paired difference, the number of folds in which the variant is worse and the $p$-values. The standard error of a mean paired difference over eight folds is about 6 veh/h, well below the seed-to-seed variability of a single training (about 16 veh/h), so that differences are judged by the paired tests and not against that floor.

### S9. Statistical comparison of the model variants

For each held-out counter and each regime, the held-out $MAE$ of the deployed configuration is divided by the standard deviation of the counter's observed flows over the same hours. Related to the variability of the counter itself, the errors of quiet streets, whose hourly counts fluctuate strongly in relative terms, and of axes become comparable. Table S6 gives, by class of mean observed flow, the median coefficient of variation of the counters, the median relative held-out $MAE$ and the median ratio of the $MAE$ to the standard deviation, with its quartiles over the counters.

**Table S6.** Held-out error of the deployed configuration normalized by the standard deviation of the counters, by class of mean observed flow (veh/h). CV: standard deviation of the observed hourly flows relative to their mean (%). Model: relative held-out $MAE$ (%). MAE/SD: held-out $MAE$ divided by the standard deviation of the counter, median and quartiles over the counters.

| Mean flow | Interpolation | | | | Extrapolation | | |
|---|---|---|---|---|---|---|---|
| | Counters | CV | Model | MAE/SD | Counters | Model | MAE/SD |
| ≤ 150 | 20 | 69 | 143 | 2.45 (1.30–3.55) | 47 | 156 | 2.51 (1.13–4.12) |
| 150–300 | 56 | 58 | 45 | 0.74 (0.54–1.05) | 112 | 47 | 0.73 (0.52–1.27) |
| 300–600 | 105 | 51 | 37 | 0.73 (0.47–0.95) | 208 | 35 | 0.66 (0.49–1.05) |
| 600–1,200 | 86 | 47 | 32 | 0.64 (0.51–1.05) | 146 | 36 | 0.78 (0.54–1.10) |
| 1,200–2,500 | 27 | 44 | 36 | 0.70 (0.55–0.98) | 53 | 42 | 0.94 (0.65–1.36) |
| > 2,500 | 49 | 42 | 13 | 0.31 (0.21–0.45) | 92 | 16 | 0.39 (0.28–0.56) |
| all | 343 | 48 | 34 | 0.64 (0.46–0.99) | 658 | 36 | 0.72 (0.49–1.14) |

## S10. Held-out error by road importance class

The held-out error of the deployed configuration is broken down by the BD TOPO importance class of the counted segment. Table S7 gives, per class and regime, the number of segments of the class in the graph, the number of counters evaluated (at least 100 held-out pairs), the median mean flow of these counters, and the medians over the counters of the relative held-out MAE, of the MAE divided by the standard deviation of the counter and of the relative bias. No counter lies on a class-6 segment. The relative error follows the flow level of the class rather than its rank, and the two classes whose counters carry less than 300 veh $h^{-1}$ are also the two that are underestimated.

**Table S7.** Held-out error of the deployed configuration by BD TOPO importance class of the counted segment, Paris intra-muros. Segments: number of segments of the class in the graph. Flow: median of the mean observed flow of the counters (veh/h). Model: median relative held-out $MAE$ (%). MAE/SD: median ratio of the $MAE$ to the standard deviation of the counter. Bias: median relative bias (%).

| Class | Segments | Interpolation | | | | | Extrapolation | | | | |
|---|---|---|---|---|---|---|---|---|---|---|---|
| | | Counters | Flow | Model | MAE/ SD | Bias | Count ers | Flow | Model | MAE/ SD | Bias |
| 1 | 831 | 61 | 3,614 | 14 | 0.34 | +1 | 115 | 3,737 | 18 | 0.46 | −5 |
| 2 | 4,082 | 61 | 301 | 43 | 0.71 | −18 | 127 | 282 | 46 | 0.78 | −22 |
| 3 | 7,824 | 180 | 606 | 37 | 0.71 | +1 | 340 | 547 | 37 | 0.78 | −5 |
| 4 | 11,411 | 28 | 327 | 37 | 0.68 | +24 | 57 | 348 | 37 | 0.69 | +14 |
| 5 | 33,566 | 13 | 240 | 55 | 0.93 | −14 | 19 | 240 | 51 | 0.92 | −14 |